\documentclass[runningheads]{llncs}
\usepackage[T1]{fontenc}
\usepackage{amssymb, amsmath}
\usepackage{booktabs}
\usepackage{multirow}
\usepackage{graphicx}
\usepackage[acronym]{glossaries}
\usepackage{hyperref}
\usepackage{cleveref}

\usepackage{color}

\newcommand{\x}{\mathbf{x}}
\newcommand{\y}{\mathbf{y}}
\newcommand{\w}{\mathbf{w}}
\newcommand{\I}{\mathbf{I}}
\newcommand{\bSigma}{\mathbf{\Sigma}}
\newcommand{\bmu}{\boldsymbol{\mu}}

\begin{document}

% Acronyms
\newacronym{APL}{APL}{Active-Passive Loss}
\newacronym{CE}{CE}{Cross Entropy}
\newacronym{CNN}{CNN}{Convolutional Neural Network}
\newacronym{EM}{EM}{Expectation Maximization}
\newacronym{GCE}{GCE}{Generalized Cross Entropy}
\newacronym{MAE}{MAE}{Mean Absolute Error}
\newacronym{MLE}{MLE}{Maximum Likelihood Estimation}
\newacronym{MSE}{MSE}{Mean Squared Error}
\newacronym{NCE}{NCE}{Normalized Cross Entropy}
\newacronym{NGCE}{NGCE}{Normalized Generalized Cross Entropy}
\newacronym{PDF}{PDF}{Probability Density Function}
\newacronym{RCE}{RCE}{Reverse Cross Entropy}
\newacronym{SCE}{SCE}{Symmetrical Cross Entropy}
\newacronym{ViT}{ViT}{Visual Transformer}

\title{Robust T-Loss for Medical Image Segmentation} 
%
% \titlerunning{T-Loss}
\author{
  Alvaro~Gonzalez-Jimenez\inst{1} \and 
  Simone~Lionetti\inst{2} \and
  Philippe~Gottfrois\inst{1} \and
  Fabian~Gr\"oger\inst{1,2} \and
  Marc~Pouly\inst{2} \and
  Alexander~Navarini\inst{1}
}

\authorrunning{Gonzalez-Jimenez et al.}

\institute{University of Basel, Switzerland \and 
Lucerne School of Computer Science and Information Technology, Switzerland
\email{alvaro.gonzalezjimenez@unibas.ch}
} 

\maketitle              % typeset the header of the contribution
\begin{abstract}
This paper presents a new robust loss function, the T-Loss, for medical image
segmentation. The proposed loss is based on the negative log-likelihood
of the Student-t distribution and can effectively handle outliers in the data
by controlling its sensitivity with a single parameter. This parameter is
updated during the backpropagation process, eliminating the need for additional 
computation or prior information about the level and spread of noisy labels.
Our experiments show that the T-Loss outperforms traditional loss functions in
terms of dice scores on two public medical datasets for skin lesion
and lung segmentation. We also demonstrate the ability of T-Loss
to handle different types of simulated label noise, resembling human error. Our results provide strong
evidence that the T-Loss is a promising alternative for medical image segmentation
where high levels of noise or outliers in the dataset are a typical phenomenon in practice.
The project website can be found at \url{https://robust-tloss.github.io}.

\end{abstract}

\section{Introduction}

% Medical image annotations
\Glspl{CNN} and \glspl{ViT}
have become the standard in semantic segmentation, achieving
state-of-the-art results in many applications
\cite{amruthalingamObjectiveHandEczema2023,shenDeepLearningMedical2017,litjensSurveyDeepLearning2017}.
However, supervised training of \glspl{CNN} and \glspl{ViT} requires
large amounts of annotated data, where each pixel in the image is labeled
with the category it belongs to.
In the medical domain, obtaining these
annotations can be costly and time-consuming as it requires expertise and
domain knowledge that is often scarcely available \cite{dlovaPrevalenceSkinDiseases2017}.
In addition, medical image annotations can be affected by human bias and poor
inter-annotator agreement \cite{ribeiroHandlingInterAnnotatorAgreement2019}, further complicating the process.
Despite efforts to obtain labels through automated mining 
\cite{yanDeepLesionAutomatedMining2018} and crowd-sourcing methods
\cite{gurariHowCollectSegmentations2015}, 
the quality of datasets gathered using these methods
remains challenging due to often high levels of label noise.

For instance, the Fitzpatrick 17k dataset, commonly used in dermatology research, contains non-skin images and noisy annotations.
In a random sample of 504 images, 5.4\% were labeled incorrectly or as other classes \cite{grohEvaluatingDeepNeural2021a}.
The dataset was scraped from online atlases, which makes it vulnerable to inaccuracies and noise \cite{grohEvaluatingDeepNeural2021a}.
Noisy labels are and will continue to be, a problem in medical datasets.
This is a concern as label noise has been shown to decrease
the accuracy of supervised models
\cite{nettletonStudyEffectDifferent2010,pechenizkiyClassNoiseSupervised2006,zhuClassNoiseVs2004},
making it a key area of focus for both research and practical applications.

% Noisy labels SotA
Previous literature has explored many methods to mitigate the problem of noisy labels in deep learning.
These methods can be broadly categorized into label correction
\cite{tongxiaoLearningMassiveNoisy2015,veitLearningNoisyLargeScale2017,yangEstimatingInstancedependentBayeslabel2022},
loss function correction based on an estimated noise transition matrix
\cite{patriniMakingDeepNeural2017,yaoDualReducingEstimation2020,wangNoiseRobustFrameworkAutomatic2020},
and robust loss functions
\cite{zhangGeneralizedCrossEntropy2018,wangSymmetricCrossEntropy2019,barron2019general,maNormalizedLossFunctions2020}.
Compared to the first two approaches, which may suffer from inaccurate estimates of the noise transition matrix,
robust loss functions enable joint optimization of model parameters and variables related to the noise model and have shown promising results in classification tasks \cite{ghoshRobustLossFunctions2017,zhangGeneralizedCrossEntropy2018}.
Despite these advances, semantic segmentation with noisy
labels is relatively understudied. Existing research in this area has focused
on the development of noise-resistant network
architectures \cite{liSuperpixelGuidedIterativeLearning2021}, the incorporation 
of domain-specific prior knowledge \cite{wangNoiseRobustFrameworkAutomatic2020}, or
more recent strategies that update the noisy masks before
memorization \cite{liuAdaptiveEarlyLearningCorrection2022}.

% Problem statement
Although previous methods have shown robustness in semantic segmentation,
they often have limitations, such as  more hyper-parameters, modifications to the network architecture, or complex training procedures.
In contrast, robust loss functions offer a much simpler solution as they could be incorporated with a simple change in a single modeling component. However,
their effectiveness has not been thoroughly investigated.

% Paper content
In this work, we show that several traditional robust loss functions
are vulnerable to memorizing noisy labels.
To overcome this problem, we introduce a novel robust loss function, the T-Loss,
which is inspired by the negative log-likelihood of the Student-t distribution.
The T-Loss, whose simplest formulation features a single parameter,
can adaptively learn an optimal tolerance level to label noise directly during backpropagation,
eliminating the need for additional computations such as the \gls{EM} steps.

% Paper content - evaluation of the T-Loss
To evaluate the effectiveness of the T-Loss as a robust loss function for medical semantic segmentation, we conducted experiments on two widely-used benchmark datasets in the field: one for skin lesion segmentation and the other for lung segmentation.
We injected different levels of noise into these datasets that simulate typical human labeling errors and trained deep learning models using various robust loss functions.
Our experiments demonstrate that the T-Loss outperforms these robust state-of-the-art loss functions in terms of segmentation accuracy and robustness, particularly under conditions of high noise contamination.
We also observed that the T-Loss could adaptively learn the optimal tolerance level to label noise which significantly reduces the risk of memorizing noisy labels.

% Paper outline
This research is divided as follows: \Cref{sec:methodology} introduces the motivation behind our T-Loss and provides its mathematical derivation. \Cref{sec:experiments} covers the datasets used in our experiments, the implementation and training details of T-Loss, and the metrics used for comparison.
\Cref{sec:results} presents the main findings of our study, including the results of the T-Loss and the baselines on both datasets and an ablation study on the parameter of T-Loss.
Finally, in \cref{sec:conclusions}, we summarize our contributions and the significance of our study for the field.

\section{Methodology}
\label{sec:methodology}

% Problem definition
Let $\x_i \in \mathbb{R}^{c \times w \times h}$ be an input image
and $\y_i \in \{0, 1\}^{w \times h}$ be its noisy annotated binary segmentation mask,
where $c$ represents the number of channels, $w$ the image's width, and $h$ its height.
Given a set of images $\{\x_1, \ldots, \x_N\}$
and corresponding masks $\{\y_1, \ldots, \y_N\}$,
our goal is to train a model $f_\w$ with parameters $\w$
such that $f_\w(\x)$ approximates
the accurate binary segmentation mask for any given image $\x$.

% Heuristics
To this end we note that, heuristically, assuming error terms to follow a Student-t distribution
(as suggested e.g.\ in \cite{murphyMachineLearningProbabilistic2012})
allows for significantly larger noise tolerance with respect to the usual gaussian form.
Recall that the Student-t distribution for a $D$-dimensional variable $\y$ is defined by the \gls{PDF}
\begin{equation}
p(\y | \bmu, \bSigma; \nu) = 
\frac{\mathrm{\Gamma}\big(\frac{\nu+D}{2}\big)}{\mathrm{\Gamma}\big(\frac{\nu}{2}\big)}
\frac{|\bSigma|^{-1/2}}{(\pi \nu)^{D/2}}
\left[
	1 + \frac{(\y-\bmu)^T \bSigma^{-1}(\y-\bmu)}{\nu}
\right]^{-\frac{\nu+D}{2}},
\end{equation}
where $\bmu$ and $\bSigma$ are respectively the mean and the covariance matrix
of the associated multivariate normal distribution,
$\nu$ is the number of degrees of freedom,
and $|\cdot|$ indicates the determinant
(see e.g.\ \cite{bishopPatternRecognitionMachine2006}).
From this expression, we see that the tails of the Student-t distribution
follow a power law that is indeed heavier compared to the usual negative quadratic exponential.
For this reason, it is well known to be robust to outliers
\cite{sunRobustMixtureClustering2010,forbesNewFamilyMultivariate2014}.

% Loss function definition
Since the common \gls{MSE} loss is derived by minimizing the negative log-likelihood of the normal distribution,
we choose to apply the same transformation and get
\begin{multline}
- \log p(\y | \bmu, \bSigma; \nu) =
- \log\mathrm{\Gamma}\Big(\frac{\nu+D}{2}\Big) + \log\mathrm{\Gamma}\Big(\frac{\nu}{2}\Big)
+ \frac{1}{2} \log |\mathbf{\Sigma}| + \frac{D}{2} \log (\pi\nu) \\
+ \frac{\nu+D}{2} \log \left[1 + \frac{(\y-\bmu)^T \bSigma^{-1}(\y-\bmu)}{\nu}\right].
\label{eq:log-student}
\end{multline}
The functional form of our loss function for one image is then obtained
with the identification $\y = \y_i$ and the approximation $\bmu = f_\w(\x_i)$,
and aggregated with
\begin{equation}
\mathcal{L}_{\text{T}} = \frac{1}{N} \sum_{i=1}^N - \log p(\y_i | f_\w(\x_i), \bSigma; \nu).
\label{eq:t_loss}
\end{equation}

% Covariance simplification
\Cref{eq:log-student} has $D(D+1)/2$ free parameters in the covariance matrix,
which should be estimated from the data.
In the case of images, this can easily be in the order of $10^4$ or larger,
which makes a general computation highly non-trivial and may deteriorate the generalization capabilities of the model.
For these reasons, we take $\bSigma$ to be the identity matrix $\I_D$,
despite knowing that pixel annotations in an image are not independent.
The loss term for one image simplifies to
\begin{multline}
- \log p(\y | \bmu, \I_D; \nu) =
- \log\mathrm{\Gamma}\Big(\frac{\nu+D}{2}\Big) + \log\mathrm{\Gamma}\Big(\frac{\nu}{2}\Big)
+ \frac{D}{2} \log (\pi\nu) \\
+ \frac{\nu+D}{2} \log \left[1 + \frac{(\y-\bmu)^2}{\nu}\right].
\label{eq:log-student-simple}
\end{multline}

% Limits
To clarify the relation with known loss functions, let $\delta = |\y_i-f_\w(\x_i)|$,
and fix the value of $\nu$.
For $\delta\to 0$, the functional dependence from $\delta$
reduces to a linear function of $\delta^2$, i.e.\ \gls{MSE}.
For large values of $\delta$, though, \cref{eq:log-student-simple}
is equivalent to $\log\delta$, thus penalizing large deviations
even less than the much-advocated robust \gls{MAE}.
The scale of this transition, the sensitivity to outliers,
is regulated by the parameter $\nu$.

% \nu
We optimize the parameter $\nu$ jointly with $\w$ using gradient descent.
To this end, we reparametrize $\nu = e^{\tilde\nu} + \epsilon$
where $\epsilon$ is a safeguard for numerical stability.
Loss functions with similar dynamic tolerance parameters
were also studied in \cite{barron2019general} in the context of regression,
where using the Student-t distribution is only mentioned in passing.

\section{Experiments}
\label{sec:experiments}

In this section, we demonstrate the robustness of the T-Loss for segmentation
tasks on two public image collections from different medical modalities,
namely ISIC \cite{codellaSkinLesionAnalysis2018} and Shenzhen \cite{Candemir2014,Jaeger2014,Stirenko2018}.
In line with the literature, we use simulated label noise
in our tests, as no public benchmark with real label noise exists \cite{liSuperpixelGuidedIterativeLearning2021}.

\subsection{Datasets}

% ISIC
The \textbf{ISIC} 2017 dataset \cite{codellaSkinLesionAnalysis2018} is a well-known public
benchmark of dermoscopy images for skin cancer detection. It contains 2000
training and 600 test images with corresponding lesion boundary masks. The
images are annotated with lesion type, diagnosis, and
anatomical location metadata. The dataset also includes a list of lesion
attributes, such as size, shape, and color. We resized the images
to $256\times256$ pixels for our experiments. 

% Shenzen
\textbf{Shenzhen} \cite{Candemir2014,Jaeger2014,Stirenko2018} is a public dataset containing 566
frontal chest radiographs with corresponding lung segmentation masks for
tuberculosis detection. Since there is not a predefined split for Shenzhen as in ISIC, to ensure representative training and testing sets,
we stratified the images by their tuberculosis and normal lung labels, with
70\% of the data for training and the remaining 30\% for testing. Resulting
in 296 training images and 170 test images. All images were resized to
$256\times256$ pixels.  

% Artificial noise
Without a public benchmark with real noisy and clean segmentation masks,
we artificially inject additional mask noise
in these two datasets to test the model's robustness to low annotation quality.
This simulates the real risk of errors due to
factors like annotator fatigue and difficulty in annotating certain images.
In particular, we follow \cite{liSuperpixelGuidedIterativeLearning2021},
randomly sample a portion of the training data with probability $\alpha \in \{ 0.3, 0.5, 0.7 \}$,
and apply morphological transformations with noise levels controlled by $\beta \in \{ 0.5, 0.7 \}$ \footnote{\url{https://github.com/gaozhitong/SP_guided_Noisy_Label_Seg}}. The morphological transformations included erosion, dilation, and affine transformations, which respectively reduced, enlarged, and displaced the annotated area.
%In figure \cref{} we show some noisy examples.

\subsection{Setup}

% Training parameters
We train a nnU--Net \cite{isenseeNnUNetSelfconfiguringMethod2021}
as a segmentation network from scratch. To increase variations in the training data,
we augment them with random mirroring, flipping, and gamma transformations.
The T-loss was initialized with $\tilde{\nu}=0$
and $\epsilon = 10^{-8}$. The nnU-Net was trained for
100 epochs using the Adam optimizer with a learning rate of $10^{-3}$
and a batch size of 16 for the ISIC dataset and 8 for the Shenzhen dataset.
The network was trained on a single NVIDIA Tesla V100 with 32 GB of memory.

% Evaluation
The model is trained using noisy masks. However, by using the ground truth 
for the corresponding noisy mask, we can evaluate the robustness of the model and measure
noisy-label memorization.
This is done by analyzing the dice score of the model's prediction compared to the actual ground truth. 

% Baselines
In addition to the T-Loss, we train several other losses for comparison.
Our analysis includes some traditional robust losses, such as \acrfull{MAE},
\acrfull{RCE}, \acrfull{NCE}, and \acrfull{NGCE}, as well as more
recent methods, such as \acrfull{GCE} \cite{zhangGeneralizedCrossEntropy2018},
\acrfull{SCE} \cite{wangSymmetricCrossEntropy2019}, and
\acrfull{APL} \cite{maNormalizedLossFunctions2020}. For
\acrshort{APL}, in particular, we consider three combinations: 1)
\acrshort{NCE}+\acrshort{RCE}, 2) \acrshort{NGCE}+\acrshort{MAE}, and 3)
\acrshort{NGCE}+\acrshort{RCE}. We consider the mean of the predictions for the last 10 epochs
with a fixed number of epochs and report its mean and standard deviation over 3 different random seeds.

% Statistical Tests
Finally, we complete our evaluation with statistical significance tests.
We use the ANOVA test \cite{girdenANOVARepeatedMeasures1992} to compare the differences between the
means of the dice scores and obtain a $\textit{p-value}$.
In addition, if the difference is significant,
we perform the Tukey $\textit{post-hoc}$ test \cite{keselmanTukeyMultipleComparison1977}
to determine which means are different.
We assume statistical significance for $p$-values of less than $p = 0.05$
and denote this with a $^\star$.

\section{Results}
\label{sec:results}

\subsection{Results on the ISIC dataset}
\begin{figure}
    \centering
    \includegraphics[width=0.87\columnwidth]{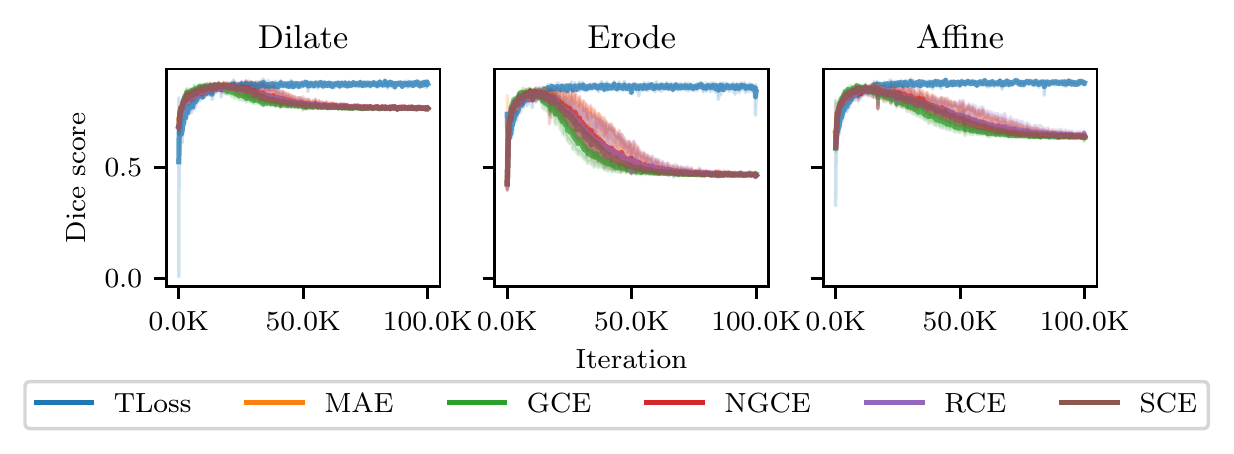}
    \caption{The dice score of training set predictions compared to
      ground truth annotations during the training process on the ISIC 2017 dataset
      for each type of noisy mask with $\alpha=0.7$, $\beta=0.7$.
      The model memorizes the noisy labels after the first $\sim$20K iterations,
      thus negatively affecting the dice score for all losses except the T-Loss.}
     \label{fig:train_isic}
\end{figure}

We present experimental results for the skin lesion segmentation task on the
ISIC dataset in \cref{tab:isic}. Our results show that conventional losses perform well with no noise or under low
noise levels, but their performance decreases significantly with increasing
noise levels due to the memorization of noisy labels.
This can be observed from the training dice scores in \cref{fig:train_isic},
where traditional robust losses overfit data in later stages of learning
while metrics for the T-Loss do not deteriorate.
Our method achieves a dice score of $0.788\pm0.007$ even for the most extreme noise scenario under exam.
Examples of the obtained masks can be seen in the supplementary material.

\begin{table}
\centering
\caption{Dice score on the ISIC dataset with different noise ratios. The values refer to the mean and standard deviation over 3 different random seeds for the mean score over the last 10 epochs.}
\label{tab:isic}
\resizebox{\textwidth}{!} {
\begin{tabular}{@{}lccccccc@{}}
\toprule
\multirow{2}{*}{Loss} & \multirow{2}{*}{$\alpha=0.0$} & \multicolumn{2}{c}{$\alpha=0.3$} & \multicolumn{2}{c}{$\alpha=0.5$} & \multicolumn{2}{c}{$\alpha=0.7$} \\
& & $\beta=0.5$ & $\beta=0.7$ & $\beta=0.5$ & $\beta=0.7$ & $\beta=0.5$ & $\beta=0.7$ \\ \midrule
GCE          & $0.828(7)$    & $0.805(9)$   & $0.785(14)$ & $0.772(15)$ & $0.736(17)$ & $0.743(12)$ & $0.691(22)$ \\
MAE          & $0.826(6)$    & $0.803(7)$   & $0.786(12)$ & $0.771(10)$ & $0.742(14)$ & $0.751(09)$ & $0.698(20)$ \\
RCE          & $0.827(5)$    & $0.802(6)$   & $0.791(11)$ & $0.779(11)$ & $0.745(17)$ & $0.752(10)$ & $0.695(16)$ \\
SCE          & $0.828(6)$    & $0.806(9)$   & $0.793(11)$ & $0.774(12)$ & $0.738(14)$ & $0.756(10)$ & $0.691(20)$ \\
NGCE         & $0.825(6)$    & $0.803(7)$   & $0.788(08)$ & $0.773(11)$ & $0.745(16)$ & $0.745(11)$ & $0.688(19)$ \\ \midrule
NCE+RCE      & $0.829(6)$    & $0.799(8)$   & $0.792(12)$ & $0.777(11)$ & $0.751(18)$ & $0.746(11)$ & $0.696(13)$ \\
NGCE+MAE     & $0.828(5)$    & $0.802(8)$   & $0.788(13)$ & $0.774(10)$ & $0.741(18)$ & $0.748(11)$ & $0.693(15)$ \\
NGCE+RCE     & $0.827(7)$    & $0.807(7)$   & $0.790(11)$ & $0.776(13)$ & $0.736(17)$ & $0.748(10)$ & $0.689(17)$ \\ \midrule
T-Loss (Ours) & $0.825(5)$ & $0.809(6)$   & $0.804(5)^\star$ & $0.800(11)^\star$ & $0.790(5)^\star$ & $0.788(7)^\star$ & $0.761(6)^\star$ \\ \bottomrule
\end{tabular}
}

\end{table}

\begin{figure}
    \centering
    \includegraphics[width=0.87\columnwidth]{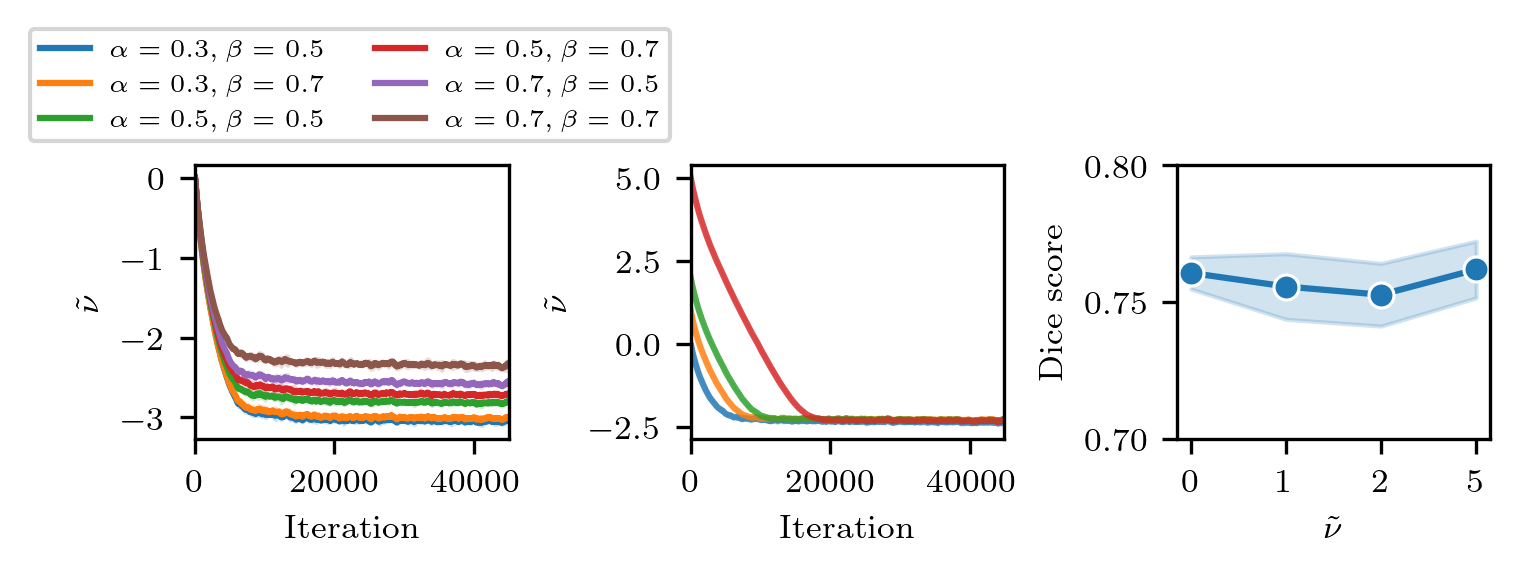}
     \caption{The behavior of $\tilde\nu$ in the skin lesion segmentation task. Left: convergence of $\tilde\nu$ with different levels of label noise. Center: sensitivity of $\tilde\nu$ to initialization for $\alpha=0.7$, $\beta=0.7$. Right: sensitivity of the dice score to the initialization of $\tilde\nu$ with the same settings.}
  \label{fig:nu-tilde}
\end{figure}

\subsection{Results on the Shenzhen dataset}
The results of lung segmentation for the Shenzhen test set are reported in 
\cref{tab:shenzhen}. Similar to the ISIC dataset, all considered robust
losses perform well at low noise levels. However, as the noise level increases,
their dice scores deteriorate. On the other hand, the T-Loss stands out by
consistently achieving the highest dice score, even in the most challenging
scenarios. The
statistical test results also support this claim, with the T-Loss
being significantly superior to the other methods.
%($p < 0.05$) for all noise
%levels, namely \acrshort{GCE} \acrshort{MAE}, \acrshort{RCE}, \acrshort{SCE}, \acrshort{NGCE}, and \acrshort{APL}.

\begin{table}
\centering
\caption{Dice score on the Shenzhen dataset with different noise ratios. The values refer to the mean and standard deviation over 3 different random seeds for the mean score over the last 10 epochs.}
\label{tab:shenzhen}
\resizebox{\textwidth}{!} {
\begin{tabular}{@{}lccccccc@{}}
\toprule
\multirow{2}{*}{Loss} & \multirow{2}{*}{$\alpha=0.0$} & \multicolumn{2}{c}{$\alpha=0.3$} & \multicolumn{2}{c}{$\alpha=0.5$} & \multicolumn{2}{c}{$\alpha=0.7$} \\
 &  & $\beta=0.5$ & $\beta=0.7$ & $\beta=0.5$ & $\beta=0.7$ & $\beta=0.5$ & $\beta=0.7$ \\ \midrule
GCE           & $0.948(1)$  & $0.933(6)$   & $0.930(8)$ & $0.909(12)$ & $0.880(19)$ & $0.856(14)$ & $0.807(23)$ \\
MAE           & $0.949(2)$  & $0.937(6)$   & $0.931(8)$ & $0.910(11)$ & $0.880(16)$ & $0.864(20)$ & $0.823(23)$ \\
RCE           & $0.949(2)$  & $0.938(4)$   & $0.931(8)$ & $0.910(10)$ & $0.886(20)$ & $0.863(15)$ & $0.818(26)$ \\
SCE           & $0.949(3)$  & $0.938(4)$   & $0.930(6)$ & $0.908(09)$ & $0.881(18)$ & $0.865(15)$ & $0.821(24)$ \\
NGCE          & $0.949(2)$  & $0.936(7)$   & $0.933(9)$ & $0.906(12)$ & $0.875(23)$ & $0.865(17)$ & $0.822(24)$ \\ \midrule
NCE+RCE       & $0.949(2)$  & $0.936(7)$   & $0.928(9)$ & $0.906(12)$ & $0.879(18)$ & $0.863(18)$ & $0.818(23)$ \\
NGCE+MAE      & $0.949(1)$  & $0.938(5)$   & $0.934(6)$ & $0.906(11)$ & $0.877(19)$ & $0.865(16)$ & $0.824(21)$ \\
NGCE+RCE      & $0.949(2)$  & $0.936(5)$   & $0.930(10)$& $0.909(10)$ & $0.884(17)$ & $0.862(12)$ & $0.821(26)$ \\ \midrule
T-Loss (Ours) & $0.949(1)$  & $0.948(1)^\star$   & $0.939(1)^\star$ & $0.914(5)^\star$  & $0.904(8)^\star$ & $0.896(7)^\star$  & $0.870(31)^\star$ \\ \bottomrule
\end{tabular}
}
\end{table}

\subsection{Dynamic tolerance to noise}

The value of $\tilde\nu$ is crucial for the model's performance,
as it controls the sensitivity to label noise.
To shed light on this mechanism, we study the behavior of $\tilde\nu$ during training for different label noise levels and initializations on the ISIC dataset.
As seen in \cref{fig:nu-tilde},
$\tilde\nu$ dynamically adjusts annotation noise tolerance in the early stages of training, independently of its initial value.
The plots demonstrate that $\tilde\nu$ clearly converges to a stable solution during training,
with initializations far from this solution only mildly prolonging the time needed for convergence
and having no significant effect on the final dice score.

\section{Conclusions}
\label{sec:conclusions}

% Methods and result summary
In this contribution, we introduced the T-Loss,
a loss function based on
the negative log-likelihood of the Student-t distribution.
The T-Loss offers the great advantage
of controlling sensitivity to outliers through a single parameter
that is dynamically optimized.
Our evaluation on public medical datasets for skin lesion and lung segmentation
demonstrates that the T-Loss outperforms other robust losses
by a statistically significant margin.
While other robust losses are vulnerable to noise memorization for high noise levels,
the T-Loss can reabsorb this form of overfitting
into the tolerance level $\nu$.
Our loss function also features remarkable independence
to different noise types and levels.

% Include limitation 
It should be noted that other methods, such as
\cite{liSuperpixelGuidedIterativeLearning2021} offer better performance
for segmentation on the ISIC dataset with the same synthetic noisy labels,
while the T-Loss offers a simple alternative.
The trade-off in terms of performance, computational cost,
and ease of adaption to different scenarios remains to be investigated.
Similarly, combinations of the T-Loss with superpixels
and/or iterative label refinement procedures are still to be explored.

% Importance of our contribution
The T-Loss provides a robust solution for binary segmentation
of medical images in the presence of high levels of annotation noise,
as frequently met in practice
e.g.\ due to annotator fatigue or inter-annotator disagreements.
This may be a key feature in achieving good generalization
in many medical image segmentation applications,
such as clinical decision support systems.
Our evaluations and analyses provide evidence that the T-Loss is a reliable
and valuable tool in the field of medical image analysis,
with the potential for broader application in other domains.

\section{Data use declaration and acknowledgment}
We declare that we have used the ISIC dataset \cite{codellaSkinLesionAnalysis2018} under the Apache License 2.0, \textcolor{blue}{\href{https://challenge.isic-archive.com/data/}{publicly available}}, and the Shenzhen dataset \cite{Candemir2014,Jaeger2014,Stirenko2018}  \textcolor{blue}{\href{https://www.kaggle.com/datasets/yoctoman/shcxr-lung-mask}{public available}} under the CC BY-NC-SA 4.0 License.

% only two pages of references
\bibliographystyle{splncs04}
\bibliography{bib}

\end{document}

\maketitle

\appendix

\section{Derivation of loss function limits}

Our final loss function is given by
\begin{equation}
\mathcal{L}_{\text{T}} = \frac{1}{N} \sum_{i=1}^N - \log p(\y_i | f_\w(\x_i), \bSigma; \nu),
\label{eq:t_loss}
\end{equation}
where
\begin{multline}
- \log p(\y_i | f_\w(\x_i), \I_D; \nu) =
- \log\mathrm{\Gamma}\Big(\frac{\nu+D}{2}\Big) + \log\mathrm{\Gamma}\Big(\frac{\nu}{2}\Big)
+ \frac{D}{2} \log (\pi\nu) \\
+ \frac{\nu+D}{2} \log \left[1 + \frac{(\y_i-f_\w(\x_i))^2}{\nu}\right].
\end{multline}
To study the behavior of the loss function for small or large errors,
we set $\delta = |\y_i-f_{\mathbf{w}}(\x_i)|$,
which leads to 
\begin{multline}
- \log p(\y_i | f_\w(\x_i), \I_D; \nu) =
- \log\mathrm{\Gamma}\Big(\frac{\nu+D}{2}\Big) + \log\mathrm{\Gamma}\Big(\frac{\nu}{2}\Big)
+ \frac{D}{2} \log (\pi\nu) \\
+ \frac{\nu+D}{2} \log \left[1 + \frac{\delta^2}{\nu}\right].
\label{eq:tlossdelta}
\end{multline}

For fixed $\nu$ and $\delta\to 0$,
all dependence on $\delta$ is in the last term and
\begin{equation}
\log \left[1 + \frac{\delta^2}{\nu}\right]
\stackrel{\delta\to 0}{\sim}
\frac{\delta^2}{\nu},
\end{equation}
so \cref{eq:tlossdelta} is a linear function of $\delta^2$.
Similarly, for fixed $\nu$ and $\delta\to+\infty$,
\begin{equation}
\log \left[1 + \frac{\delta^2}{\nu}\right]
\stackrel{\delta\to+\infty}{\sim}
\log \left[\frac{\delta^2}{\nu}\right]
=
2\log\delta - \log\nu,
\end{equation}
so \cref{eq:tlossdelta} is a linear function of $\log\delta$.

%Moreover, for fixed $\delta$ we have
%\begin{align}
%- \log p(\y_i | f_{\mathbf{w}}(\x_i), \I_D; \nu) & \stackrel{\nu\to 0}{\sim}
%D\log\delta-\log\nu+\frac{1}{2}D\log\pi-\log\mathrm{\Gamma}\Big[\frac{D}{2}\Big]+\log2, \\
%- \log p(\y_i | f_{\mathbf{w}}(\x_i), \I_D; \nu) & \stackrel{\nu\to\infty}{\sim}
%\frac{\delta^2}{2}+\frac{D}{2} \log2\pi.
%\end{align}
%The limit for $\nu\to\infty$ is therefore well-behaved,
%reducing to a linear function of $\delta^2$,
%while the limit $\nu\to 0$ has a cost which is ever increasing,
%therefore preventing the model tolerance to increase indefinitely.

\newpage

\section{Visualization of segmentation masks}

\begin{figure}[h!]
  \begin{center}
    \includegraphics[width=\textwidth]{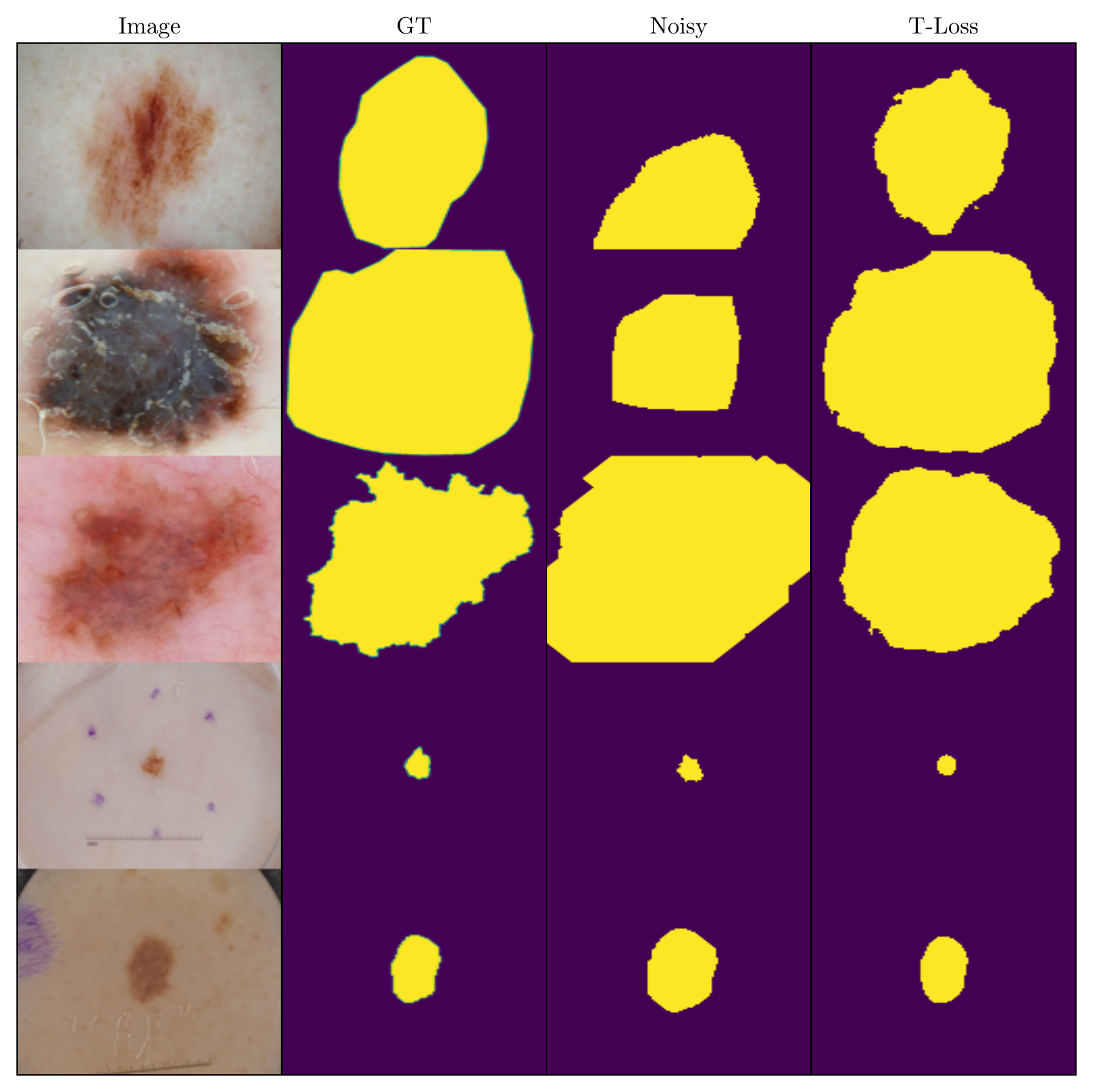}
  \end{center}
  \caption{Example of image segmentation with T-Loss on the ISIC dataset: images, ground truth masks, noisy masks, and T-Loss predictions.}
  \label{fig:masks_isic}
\end{figure}